%% file: main.tex
\documentclass[11pt]{article}
\usepackage[utf8]{inputenc}
\usepackage[T1]{fontenc}
\usepackage{lmodern}
\usepackage{amsmath,amssymb}
\usepackage{booktabs}
\usepackage{longtable}
\usepackage{array}
\usepackage{calc}
\usepackage{float}
\usepackage{graphicx}
\usepackage{caption}
\usepackage{xurl}
\usepackage{hyperref}
\usepackage[margin=1in]{geometry}
\usepackage{titlesec}
\titleformat{\section}{\Large\bfseries}{\thesection}{0.75em}{}
\titleformat{\subsection}{\large\bfseries}{\thesubsection}{0.75em}{}
\titleformat{\subsubsection}{\normalsize\bfseries}{\thesubsubsection}{0.75em}{}
\titlespacing*{\section}{0pt}{1.4em}{0.6em}
\titlespacing*{\subsection}{0pt}{1.1em}{0.4em}
\titlespacing*{\subsubsection}{0pt}{0.9em}{0.3em}
\makeatletter
\def\maxwidth{\ifdim\Gin@nat@width>\linewidth\linewidth\else\Gin@nat@width\fi}
\def\maxheight{\ifdim\Gin@nat@height>0.85\textheight0.85\textheight\else\Gin@nat@height\fi}
\makeatother
\setkeys{Gin}{width=\maxwidth,height=\maxheight,keepaspectratio}
\makeatletter
\newsavebox\pandoc@box
\newcommand*\pandocbounded[1]{%
  \sbox\pandoc@box{#1}%
  \Gscale@div\@tempa{\textheight}{\dimexpr\ht\pandoc@box+\dp\pandoc@box\relax}%
  \Gscale@div\@tempb{\linewidth}{\wd\pandoc@box}%
  \ifdim\@tempb\p@<\@tempa\p@
    \let\@tempa\@tempb
  \fi
  \ifdim\@tempa\p@<\p@
    \scalebox{\@tempa}{\usebox\pandoc@box}%
  \else
    \usebox\pandoc@box
  \fi
}
\makeatother

\title{AI Mental Models: Learned Intuition and Deliberation in a Bounded Neural Architecture}
\author{Laurence Anthony\\School of Creative Science and Engineering\\Faculty of Science and Engineering\\Waseda University\\\texttt{anthony@waseda.jp}}
\date{}

\begin{document}
\maketitle
\begin{abstract}
\input{abstract.tex}
\end{abstract}
\input{paper_body.tex}

\end{document}

%% file: abstract.tex
This paper asks whether a bounded neural architecture can exhibit a meaningful division of labor between intuition and deliberation on a classic 64-item syllogistic reasoning benchmark. More broadly, the benchmark is relevant to ongoing debates about world models and multi-stage reasoning in AI. It provides a controlled setting for testing whether a learned system can develop structured internal computation rather than only one-shot associative prediction. Experiment 1 evaluates a direct neural baseline for predicting full 9-way human response distributions under 5-fold cross-validation. Experiment 2 introduces a bounded dual-path architecture with separate intuition and deliberation pathways, motivated by computational mental-model theory (Khemlani \& Johnson-Laird, 2022). Under cross-validation, bounded intuition reaches an aggregate correlation of \texttt{r\ =\ 0.7272}, whereas bounded deliberation reaches \texttt{r\ =\ 0.8152}, and the deliberation advantage is significant across folds (\texttt{p\ =\ 0.0101}). The largest held-out gains occur for \texttt{NVC}, \texttt{Eca}, and \texttt{Oca}, suggesting improved handling of rejection responses and \texttt{c-a} conclusions. A canonical \texttt{80:20} interpretability run and a five-seed stability sweep further indicate that the deliberation pathway develops sparse, differentiated internal structure, including an \texttt{Oac}-leaning state, a dominant workhorse state, and several weakly used or unused states whose exact indices vary across runs. These findings are consistent with reasoning-like internal organization under bounded conditions, while stopping short of any claim that the model reproduces full sequential processes of model construction, counterexample search, and conclusion revision.

%% file: paper_body.tex
\section{Introduction}\label{introduction}

Syllogistic reasoning remains one of the most compact and well-studied benchmarks for theories of human reasoning (Johnson-Laird \& Steedman, 1978; Khemlani \& Johnson-Laird, 2012). The benchmark is attractive because the task space is small enough to examine exhaustively, yet rich enough to expose systematic differences among competing theories. The 64 classic syllogisms also support a stronger target than a simple correct-versus-incorrect label; human participants distribute their responses across multiple possible conclusions and \texttt{NVC} (``no valid conclusion''), yielding a structured behavioral matrix rather than a single score.

The present work begins from a simple question: what would it take for a neural architecture to model this benchmark in a way that is interesting from the perspective of cognitive theory rather than merely pattern fitting? Preliminary internal experiments in this project, including in-sample fits on simpler reasoning tasks and early syllogistic runs, suggested that high training performance can be achieved without demonstrating strong held-out generalization. The present paper therefore focuses on the harder question of whether a bounded architecture can learn reusable structure rather than only statistical regularities in the training data, and whether any such gain is accompanied by interpretable internal differentiation.

This question is relevant beyond cognitive psychology. In contemporary AI, there is increasing interest in systems that move beyond one-shot pattern completion toward richer internal organization, persistent representations, and multi-stage computation. The present paper uses syllogistic reasoning as a tightly controlled testbed for a narrower but related question: can a small learned architecture develop internally differentiated computation that improves held-out behavior in cases where a single-pass mapping is limited? A deliberate advantage of starting at this scale is that the resulting system is fully interpretable: every deliberative state, every gate weight, and every ablation outcome can be inspected directly. The question this opens is whether the architectural principles established here extend to richer domains.

Our first contribution is therefore methodological. In Experiment 1, we evaluate a direct neural baseline on the full 9-way syllogistic response-distribution task under held-out evaluation and 5-fold cross-validation. This establishes a realistic baseline against which more theory-driven architectures can be compared.

Our second contribution is architectural. In Experiment 2, we revisit the benchmark with a deliberately bounded dual-path model. A weak intuition pathway, intended to approximate a fast and capacity-limited first-pass process, is paired with a separate deliberation pathway that receives the same structured input but computes through a gated set of five deliberative states. This design is inspired by the distinction, emphasized in computational mental-model theory, between a fast initial interpretation and a more deliberate search over alternative mental models and counterexamples. The central question is whether such a deliberation pathway can improve over a weak intuition pathway on held-out syllogisms, and whether that improvement falls on theoretically meaningful parts of the response space.

The results reported below show that, under matched 5-fold cross-validation, the deliberation pathway outperforms bounded intuition, especially on \texttt{NVC} and \texttt{c-a} response categories, and that this improvement is accompanied by differentiated internal organization rather than uniform use of the deliberative mechanism. In the interpretability analyses, the deliberative states also separate into more specific roles, including an \texttt{Oac}-leaning state, a universal-conclusion contributor for \texttt{AA}/\texttt{AE} items, an \texttt{E}-premise contributor, and a recurrent pattern of sparse state use in which only a subset of the five states carry most of the computation. At the same time, the evidence does not warrant a strong claim that the network has been shown to reason in the full human sense. The paper therefore makes a narrower claim: bounded neural architectures can exhibit a useful division of labor between intuition and deliberation on a classic reasoning benchmark.

\section{Related Work}\label{related-work}

\subsection{Mental models, deliberation, and syllogisms}\label{mental-models-deliberation-and-syllogisms}

The mental-model tradition is founded on the idea that reasoning depends on the construction and revision of internal models of possible situations rather than solely on the application of explicit formal rules (Johnson-Laird \& Ragni, 2024; Khemlani \& Johnson-Laird, 2022). Within this framework, cognition proceeds not only by manipulating abstract propositions, but also by building structured internal representations of how the world, or a possible world, might be organized.

This perspective is relevant to current AI discussions as well. Recent debates about the limitations of surface pattern matching, and the corresponding need for richer internal world models, make the mental-model tradition newly interesting outside cognitive psychology. The present paper does not claim to implement a full world-model architecture in the contemporary AI sense. However, it does ask a related question: can a bounded neural architecture show evidence of internal organization that is more consistent with model construction and revision than with one-shot associative mapping alone?

Syllogisms are useful in this context because they provide a compact testbed for studying whether a model can move beyond one-shot associative mapping. Johnson-Laird and Steedman (1978) helped establish the modern item-by-item study of syllogistic reasoning. The 2012 meta-analysis by Khemlani and Johnson-Laird (2012) developed this benchmark further by assembling a full 64-syllogism response-distribution dataset and by showing that no existing theory fully captures the complete response space. We therefore use syllogisms here not as an end in themselves, but as a tractable way to investigate broader questions about internal structure, bounded intuition, and deliberation.

\subsection{Mental-model theory and mReasoner}\label{mental-model-theory-and-mreasoner}

Our main theoretical anchor is the 2022 computational mental-model paper, \emph{Reasoning About Properties: A Computational Theory} (Khemlani \& Johnson-Laird, 2022). In that account, quantified reasoning is modeled in terms of the construction of small mental models of possibilities, followed by inferences drawn over those possible situations. A key feature of the theory is its distinction between an initial, heuristic, single-model process and a more deliberate search for alternative mental models and counterexamples. Khemlani and Johnson-Laird operationalize this distinction in the \texttt{mReasoner} computer program. In the present paper, \texttt{mReasoner} serves as the main theoretical source of inspiration and comparison.

The present paper does not attempt to reproduce \texttt{mReasoner} symbolically. Instead, it asks whether a bounded neural architecture can approximate some of the same intuition-versus-deliberation division of labor while remaining learnable from data. The relationship is therefore one of theoretical inspiration rather than direct implementation. Earlier work by Bucciarelli and Johnson-Laird (1999) is also relevant here, because it showed that syllogistic reasoners appear to use multiple strategies, including forms of refutation and reconsideration that fit naturally with a distinction between first-pass and deliberative processing. This makes it important to ask whether the present model also shows evidence of differentiated internal strategy-like behavior. At the same time, the present architecture differs from \texttt{mReasoner} in an important respect: its deliberation pathway is a learned multi-state computation, not a sequential symbolic process that explicitly constructs alternative models, searches for counterexamples, and revises a conclusion.

Mental-model theory is not the only computational approach to this benchmark. Chater and Oaksford's (1999) probability heuristics model argued that much syllogistic behavior can be captured through probabilistic heuristics rather than explicit model construction. Hattori (2016) proposed a hybrid account that integrates heuristic and model-based ideas. Tessler et al.~(2022) developed a probabilistic-pragmatic account of full syllogistic response distributions. These alternative models matter here because they identify different weaknesses in purely mental-model explanations, including insufficient probabilistic treatment, weak handling of graded response tendencies, and limited attention to pragmatic interpretation. The present paper does not directly solve all of those concerns, but it does address one of them: whether a learned architecture can exhibit stronger distribution-level fit and internal differentiation without collapsing into pure in-sample pattern matching.

The intuition-versus-deliberation distinction also connects to a broader dual-process literature. Evans (2008) provides a standard overview of fast and slow reasoning accounts, while De Neys (2014, 2023) argues for a more careful treatment of what is and is not captured by such distinctions. These cautions are relevant here. The present paper uses ``intuition'' and ``deliberation'' as architectural labels motivated by the cognitive literature, but it does not assume that a better-fitting deliberative pathway has thereby been shown to instantiate the full content of dual-process theory.

\subsection{Computational-modeling criteria}\label{computational-modeling-criteria}

An important requirement for the present paper is a defensible way to evaluate a computational model of reasoning. Cassimatis et al.~(2008) argue that higher-order cognitive models should be evaluated not only in terms of fit, but also in terms of breadth and parsimony. That perspective aligns well with the goals of the present paper. A near-perfect in-sample fit would not, by itself, be persuasive. For this reason, the paper emphasizes held-out performance, cross-validation, and the internal differentiation of the deliberative mechanism rather than raw training accuracy alone.

\section{Experimental Setup}\label{experimental-setup}

\subsection{Human dataset}\label{human-dataset}

Both experiments use a machine-readable extraction of Table 6 from the 2012 meta-analysis (Khemlani \& Johnson-Laird, 2012). That extraction preserves the 64 canonical syllogisms, the associated premise forms, and the full nine-way human response distributions used as targets in the present study.

Each syllogism is paired with a human response distribution over nine response types. Table 1 lists the response labels used throughout the paper.

\begin{table}[H]
\centering
\small
\caption{Response categories used in the human target distribution.}
\begin{tabular}{@{}ll@{}}
\toprule\noalign{}
Label & Reading \\
\midrule\noalign{}
\texttt{Aac} & All \texttt{a} are \texttt{c} \\
\texttt{Eac} & No \texttt{a} are \texttt{c} \\
\texttt{Iac} & Some \texttt{a} are \texttt{c} \\
\texttt{Oac} & Some \texttt{a} are not \texttt{c} \\
\texttt{Aca} & All \texttt{c} are \texttt{a} \\
\texttt{Eca} & No \texttt{c} are \texttt{a} \\
\texttt{Ica} & Some \texttt{c} are \texttt{a} \\
\texttt{Oca} & Some \texttt{c} are not \texttt{a} \\
\texttt{NVC} & No valid conclusion \\
\bottomrule
\end{tabular}
\end{table}

Targets are represented as probability distributions obtained by dividing the reported percentages by \texttt{100}.

\subsection{Input representation}\label{input-representation}

Both experiments use the same structured multi-hot input encoding. Each syllogism is represented by a \texttt{29}-dimensional vector containing separate feature blocks for \texttt{premise\_1} and \texttt{premise\_2}. The model is therefore given explicit positional information about the two premises without adding extra derived features such as figure or mood. Those structural relations are implicit in the premise encoding itself.

\subsection{Model overview}\label{model-overview}

Experiment 1 uses a direct multilayer perceptron baseline with a \texttt{29}-dimensional input, a single \texttt{64}-unit hidden layer with \texttt{ReLU}, and a \texttt{9}-way output distribution. Experiment 2 uses a bounded dual-path model with a \texttt{4}-unit intuition state and a separate deliberation pathway consisting of a \texttt{4}-unit deliberation state, five \texttt{4}-unit candidate states, and a learned soft gate over those five candidate states. The key design feature is that the deliberation pathway does not receive the intuition hidden state as input. Both pathways read the same syllogism encoding but compute separately.

Table 2 summarizes the two model families and their core architectural differences. The direct MLP uses a single hidden layer with \texttt{ReLU}. The intuition pathway is deliberately capacity-limited, whereas the deliberation pathway uses a separate encoder, five candidate state heads, and a learned gate. Notably, the full Experiment 2 model still has many fewer trainable parameters than the direct MLP baseline. Figure 1 illustrates the bounded intuition-and-deliberation architecture used in Experiment 2, including the separate intuition and deliberation pathways, the five deliberative states, and the learned deliberation gate.

\begin{table}[H]
\centering
\small
\caption{Summary of the model variants used in the paper.}
\begin{tabular}{@{}
  >{\raggedright\arraybackslash}p{(\linewidth - 10\tabcolsep) * \real{0.1364}}
  >{\raggedright\arraybackslash}p{(\linewidth - 10\tabcolsep) * \real{0.1364}}
  >{\raggedleft\arraybackslash}p{(\linewidth - 10\tabcolsep) * \real{0.1818}}
  >{\raggedleft\arraybackslash}p{(\linewidth - 10\tabcolsep) * \real{0.1818}}
  >{\raggedleft\arraybackslash}p{(\linewidth - 10\tabcolsep) * \real{0.1818}}
  >{\raggedleft\arraybackslash}p{(\linewidth - 10\tabcolsep) * \real{0.1818}}@{}}
\toprule\noalign{}
\begin{minipage}[b]{\linewidth}\raggedright
Model
\end{minipage} & \begin{minipage}[b]{\linewidth}\raggedright
Role in paper
\end{minipage} & \begin{minipage}[b]{\linewidth}\raggedleft
Input
\end{minipage} & \begin{minipage}[b]{\linewidth}\raggedleft
Hidden / state size
\end{minipage} & \begin{minipage}[b]{\linewidth}\raggedleft
Output
\end{minipage} & \begin{minipage}[b]{\linewidth}\raggedleft
Parameters
\end{minipage} \\
\midrule\noalign{}
Direct MLP & Experiment 1 baseline & 29 & 64 & 9-way softmax distribution & 2,505 \\
Intuition pathway & Experiment 2 first-pass system & 29 & 4 & 9-way softmax distribution & 165 \\
Deliberation pathway & Experiment 2 second-stage system & 29 & 24 & 9-way softmax distribution & 470 \\
Experiment 2 total & Combined bounded model & 29 & 4 + 24 & 2 x 9-way softmax distributions & 635 \\
\bottomrule
\end{tabular}
\end{table}

\emph{Note. Parameter counts include biases. The \texttt{24}-d deliberation representation comprises five \texttt{4}-d candidate states plus a \texttt{4}-d deliberation state.}

\begin{figure}
\centering
\pandocbounded{\includegraphics[keepaspectratio]{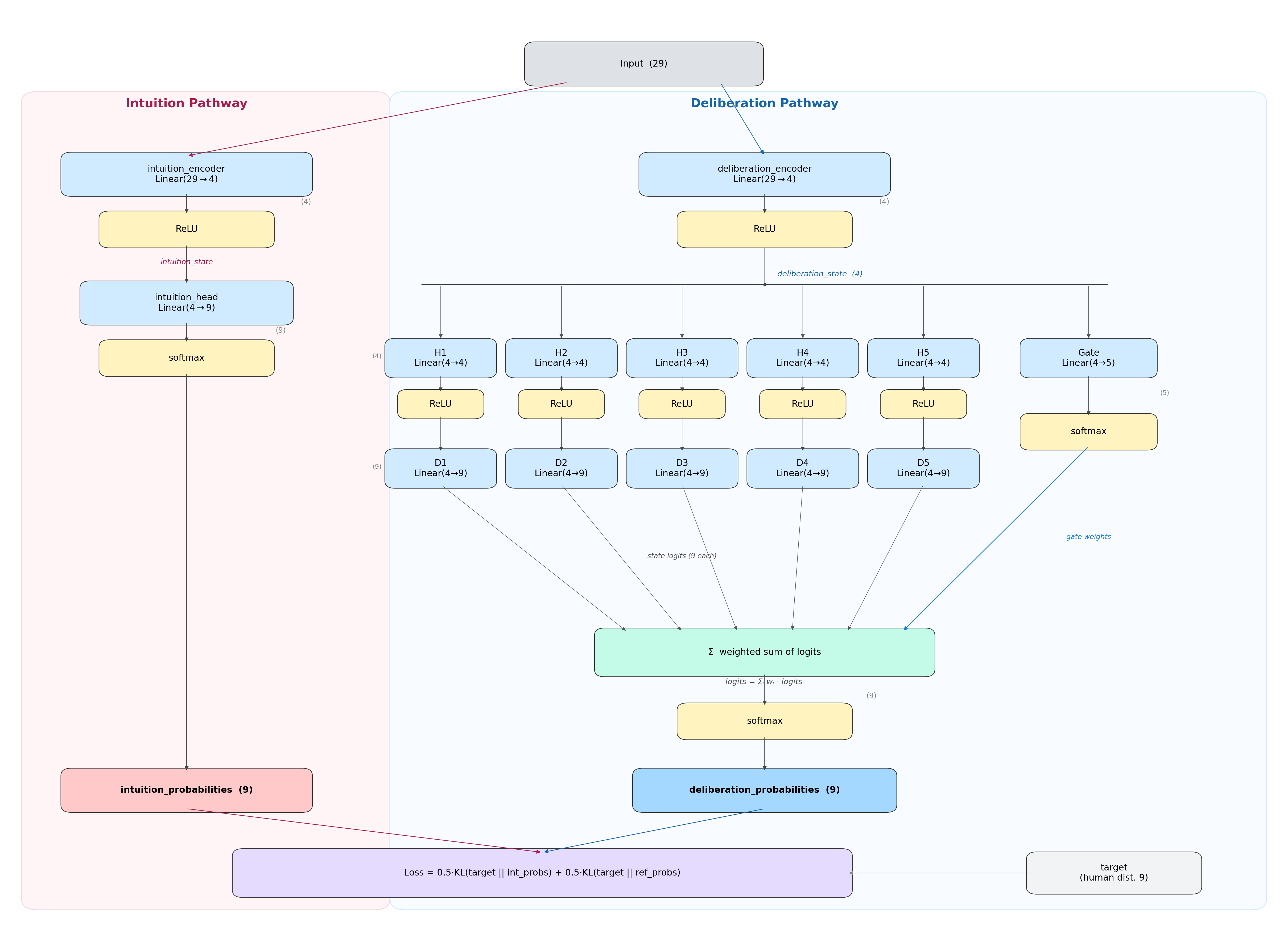}}
\caption{Architecture of the bounded intuition-and-deliberation model used in Experiment 2. The intuition pathway and the deliberation pathway receive the same structured syllogism input through separate encoders, and the deliberation pathway computes through five candidate deliberative states combined by a learned gate.}
\end{figure}

\subsection{Training}\label{training}

Both experiments are trained with the same optimizer settings. The models minimize Kullback-Leibler divergence between the predicted 9-way softmax distribution and the human target distribution, using the Adam optimizer with learning rate \texttt{1e-3} for \texttt{4000} epochs. In Experiment 2, the intuition and deliberation pathways are both trained against the same human target distribution, and their losses are combined with equal weights (\texttt{0.5} and \texttt{0.5}).

\subsection{Evaluation}\label{evaluation}

The primary evaluation reported in this paper is 5-fold cross-validation with a fixed seed. We report aggregate correlation and RMSE across the full \texttt{64\ x\ 9} response matrix, and we report fold-wise mean correlations and standard deviations as an uncertainty summary. We also report descriptive \texttt{95\%} bootstrap confidence intervals for the aggregate correlations by resampling syllogism rows while keeping their full 9-way response vectors together. The intervals use \texttt{5000} bootstrap resamples and the percentile method (\texttt{2.5th} and \texttt{97.5th} percentiles). Fold-wise correlations are used in paired statistical tests, and the bootstrap intervals are intended to complement rather than replace those tests. Single-split results are included only for qualitative illustration and interpretability analyses.

\section{Experiment 1: Direct Distribution Prediction}\label{experiment-1-direct-distribution-prediction}

Experiment 1 asks how far a direct neural baseline can go on the syllogistic response-distribution task when evaluated properly. This is important because a strong in-sample fit, by itself, is not enough to motivate a richer cognitive interpretation. Under 5-fold cross-validation, the direct MLP reaches an aggregate correlation of \texttt{r\ =\ 0.7105} with an aggregate RMSE of \texttt{0.1301}. Its mean fold correlation is \texttt{0.7140} with a standard deviation of \texttt{0.0980}. The aggregate results are summarized in Table 3, and the per-response-type breakdown is given in Table 4.

\begin{table}[H]
\centering
\small
\caption{Main 5-fold cross-validation results. The deliberation pathway shows the strongest overall held-out performance.}
\begin{tabular}{@{}
  >{\raggedright\arraybackslash}p{(\linewidth - 12\tabcolsep) * \real{0.1111}}
  >{\raggedleft\arraybackslash}p{(\linewidth - 12\tabcolsep) * \real{0.1481}}
  >{\raggedleft\arraybackslash}p{(\linewidth - 12\tabcolsep) * \real{0.1481}}
  >{\raggedleft\arraybackslash}p{(\linewidth - 12\tabcolsep) * \real{0.1481}}
  >{\raggedleft\arraybackslash}p{(\linewidth - 12\tabcolsep) * \real{0.1481}}
  >{\raggedleft\arraybackslash}p{(\linewidth - 12\tabcolsep) * \real{0.1481}}
  >{\raggedleft\arraybackslash}p{(\linewidth - 12\tabcolsep) * \real{0.1481}}@{}}
\toprule\noalign{}
\begin{minipage}[b]{\linewidth}\raggedright
Model
\end{minipage} & \begin{minipage}[b]{\linewidth}\raggedleft
Aggregate correlation
\end{minipage} & \begin{minipage}[b]{\linewidth}\raggedleft
95\% bootstrap CI
\end{minipage} & \begin{minipage}[b]{\linewidth}\raggedleft
Aggregate RMSE
\end{minipage} & \begin{minipage}[b]{\linewidth}\raggedleft
Aggregate MAE
\end{minipage} & \begin{minipage}[b]{\linewidth}\raggedleft
Mean fold correlation
\end{minipage} & \begin{minipage}[b]{\linewidth}\raggedleft
Fold SD
\end{minipage} \\
\midrule\noalign{}
Experiment 1 direct MLP & 0.7105 & {[}0.6309, 0.7825{]} & 0.1301 & 0.0675 & 0.7140 & 0.0980 \\
Experiment 2 intuition & 0.7272 & {[}0.6649, 0.7868{]} & 0.1156 & 0.0674 & 0.7212 & 0.0838 \\
Experiment 2 deliberation & 0.8152 & {[}0.7608, 0.8634{]} & 0.1029 & 0.0536 & 0.8142 & 0.0710 \\
\bottomrule
\end{tabular}
\end{table}

\begin{table}[H]
\centering
\small
\caption{Aggregate per-response-type correlations under 5-fold cross-validation.}
\begin{tabular}{@{}
  >{\raggedright\arraybackslash}p{(\linewidth - 8\tabcolsep) * \real{0.1579}}
  >{\raggedleft\arraybackslash}p{(\linewidth - 8\tabcolsep) * \real{0.2105}}
  >{\raggedleft\arraybackslash}p{(\linewidth - 8\tabcolsep) * \real{0.2105}}
  >{\raggedleft\arraybackslash}p{(\linewidth - 8\tabcolsep) * \real{0.2105}}
  >{\raggedleft\arraybackslash}p{(\linewidth - 8\tabcolsep) * \real{0.2105}}@{}}
\toprule\noalign{}
\begin{minipage}[b]{\linewidth}\raggedright
Response type
\end{minipage} & \begin{minipage}[b]{\linewidth}\raggedleft
Direct MLP
\end{minipage} & \begin{minipage}[b]{\linewidth}\raggedleft
Intuition
\end{minipage} & \begin{minipage}[b]{\linewidth}\raggedleft
Deliberation
\end{minipage} & \begin{minipage}[b]{\linewidth}\raggedleft
Deliberation - Intuition
\end{minipage} \\
\midrule\noalign{}
\texttt{Aac} & 0.9718 & 0.7353 & 0.8450 & +0.1097 \\
\texttt{Eac} & 0.7816 & 0.8259 & 0.8156 & -0.0103 \\
\texttt{Iac} & 0.7741 & 0.6894 & 0.8441 & +0.1547 \\
\texttt{Oac} & 0.8033 & 0.6336 & 0.7050 & +0.0714 \\
\texttt{Aca} & -0.0170 & 0.1682 & 0.2169 & +0.0488 \\
\texttt{Eca} & 0.4050 & 0.5203 & 0.7790 & +0.2587 \\
\texttt{Ica} & 0.4489 & 0.7396 & 0.8336 & +0.0940 \\
\texttt{Oca} & 0.4363 & 0.4235 & 0.5866 & +0.1630 \\
\texttt{NVC} & 0.4636 & 0.4443 & 0.7100 & +0.2657 \\
\bottomrule
\end{tabular}
\end{table}

These results support two conclusions. First, the benchmark is not trivial under held-out evaluation. The direct model performs meaningfully above chance-like behavior, with a \texttt{95\%} bootstrap interval of \texttt{{[}0.6309,\ 0.7825{]}} around its aggregate correlation, while remaining well below the near-perfect in-sample fits that appeared possible in preliminary internal runs before held-out evaluation was adopted. Second, the direct model remains a serious baseline. Any more theory-driven architecture should therefore be judged not only against intuition alone, but also against this direct MLP.

Qualitatively, the representative held-out heatmap in Figure 2 shows that the direct model captures broad structure in the human matrix, but still leaves substantial residual error, especially in regions associated with lower-frequency or more heterogeneous human responses. This makes Experiment 1 useful as a baseline precisely because it does \emph{not} collapse the task into either failure or trivial success.

\begin{figure}
\centering
\pandocbounded{\includegraphics[keepaspectratio]{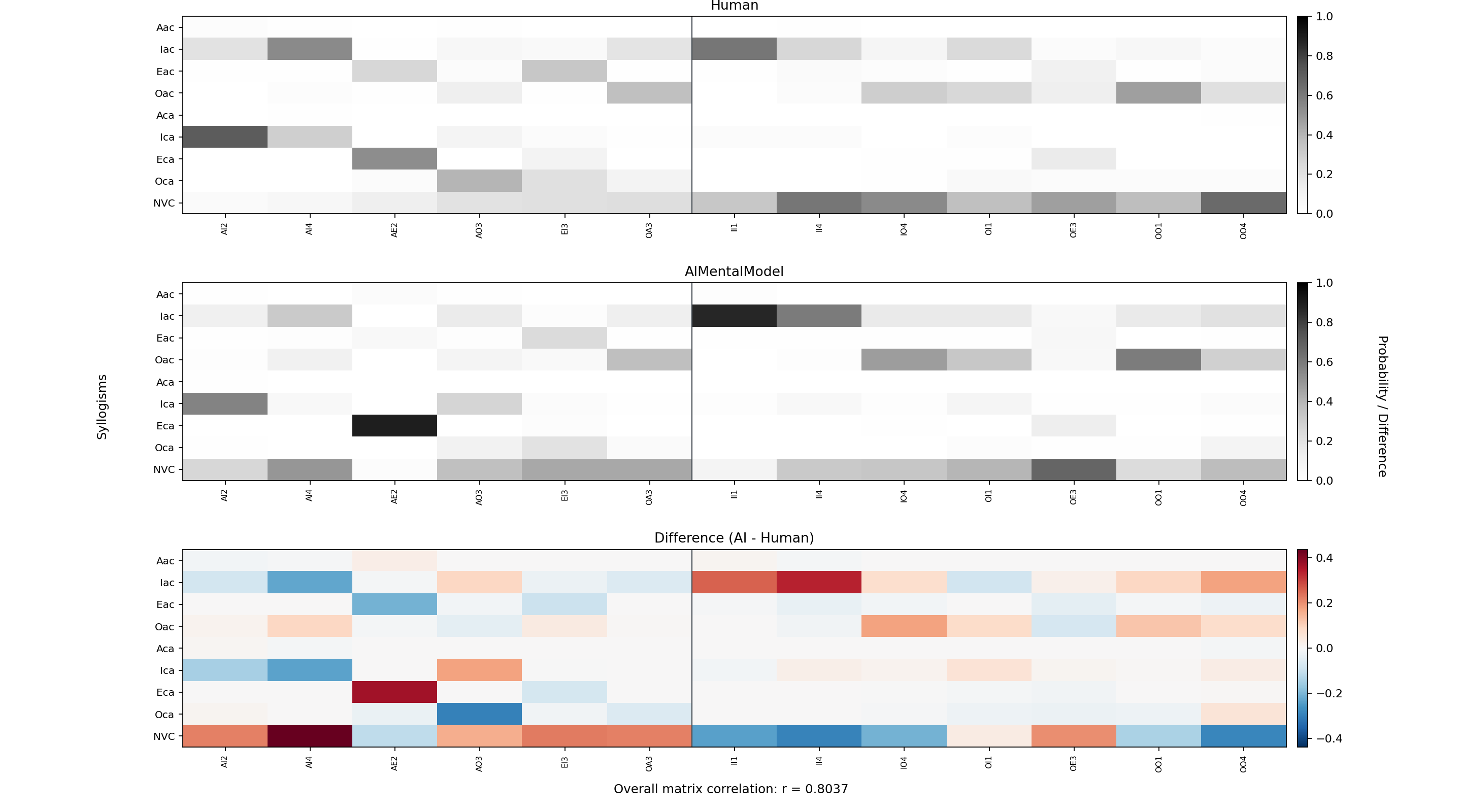}}
\caption{Representative held-out performance for the direct baseline in Experiment 1. The three panels show the human response matrix, the model prediction matrix, and the signed difference matrix on the test split.}
\end{figure}

\section{Experiment 2: Bounded Intuition and Deliberation}\label{experiment-2-bounded-intuition-and-deliberation}

Experiment 2 introduces the paper's main architectural idea: a deliberately weak intuition pathway paired with a separate bounded deliberation pathway. The aim is not to maximize raw fit with a large model, but to create a regime in which deliberation has room to matter.

Table 3 summarizes the aggregate results, Table 4 gives the per-response-type breakdown, and Tables 5 and 6 report the paired statistical comparisons. Under 5-fold cross-validation, the intuition pathway reaches an aggregate correlation of \texttt{r\ =\ 0.7272} with aggregate RMSE \texttt{0.1156}, and a \texttt{95\%} bootstrap interval of \texttt{{[}0.6649,\ 0.7868{]}}. The deliberation pathway reaches \texttt{r\ =\ 0.8152} with aggregate RMSE \texttt{0.1029}, and its corresponding \texttt{95\%} bootstrap interval is \texttt{{[}0.7608,\ 0.8634{]}}. The 2022 paper reports that the best-fitting published \texttt{mReasoner} simulation reaches approximately \texttt{r\ =\ .80} on the full syllogistic benchmark (Khemlani \& Johnson-Laird, 2022, Figure 2). The present deliberation result is therefore numerically comparable to that published aggregate fit, though the two evaluations are not directly comparable because the present paper emphasizes held-out cross-validation whereas the published \texttt{mReasoner} fit is reported on the full benchmark. A direct item-by-item comparison with \texttt{mReasoner} is an important next step.

\begin{table}[H]
\centering
\small
\caption{Paired 5-fold statistical comparisons using fold-wise correlations.}
\begin{tabular}{@{}
  >{\raggedright\arraybackslash}p{(\linewidth - 6\tabcolsep) * \real{0.2143}}
  >{\raggedleft\arraybackslash}p{(\linewidth - 6\tabcolsep) * \real{0.2857}}
  >{\raggedleft\arraybackslash}p{(\linewidth - 6\tabcolsep) * \real{0.2857}}
  >{\raggedright\arraybackslash}p{(\linewidth - 6\tabcolsep) * \real{0.2143}}@{}}
\toprule\noalign{}
\begin{minipage}[b]{\linewidth}\raggedright
Comparison
\end{minipage} & \begin{minipage}[b]{\linewidth}\raggedleft
t
\end{minipage} & \begin{minipage}[b]{\linewidth}\raggedleft
p
\end{minipage} & \begin{minipage}[b]{\linewidth}\raggedright
Interpretation
\end{minipage} \\
\midrule\noalign{}
Deliberation vs intuition & 4.5914 & 0.0101 & Significant deliberation advantage \\
Deliberation vs direct MLP & 1.5678 & 0.1920 & Numerical advantage, not significant \\
Intuition vs direct MLP & 0.1047 & 0.9217 & No reliable difference \\
\bottomrule
\end{tabular}
\end{table}

\begin{table}[H]
\centering
\small
\caption{Fold-wise correlations used in the paired statistical comparisons.}
\begin{tabular}{@{}
  >{\raggedright\arraybackslash}p{(\linewidth - 6\tabcolsep) * \real{0.2000}}
  >{\raggedleft\arraybackslash}p{(\linewidth - 6\tabcolsep) * \real{0.2667}}
  >{\raggedleft\arraybackslash}p{(\linewidth - 6\tabcolsep) * \real{0.2667}}
  >{\raggedleft\arraybackslash}p{(\linewidth - 6\tabcolsep) * \real{0.2667}}@{}}
\toprule\noalign{}
\begin{minipage}[b]{\linewidth}\raggedright
Fold
\end{minipage} & \begin{minipage}[b]{\linewidth}\raggedleft
Experiment 1 direct MLP
\end{minipage} & \begin{minipage}[b]{\linewidth}\raggedleft
Experiment 2 intuition
\end{minipage} & \begin{minipage}[b]{\linewidth}\raggedleft
Experiment 2 deliberation
\end{minipage} \\
\midrule\noalign{}
1 & 0.8523 & 0.6599 & 0.7313 \\
2 & 0.6223 & 0.6271 & 0.7478 \\
3 & 0.5859 & 0.8284 & 0.8627 \\
4 & 0.7350 & 0.7391 & 0.8256 \\
5 & 0.7745 & 0.7517 & 0.9036 \\
\bottomrule
\end{tabular}
\end{table}

The central inferential result is that deliberation significantly outperforms intuition under matched folds (\texttt{t\ =\ 4.5914}, \texttt{p\ =\ 0.0101}, paired two-tailed test). By contrast, intuition does not differ meaningfully from the direct MLP baseline (\texttt{t\ =\ 0.1047}, \texttt{p\ =\ 0.9217}), and the deliberation pathway's advantage over the direct MLP is numerically positive but not statistically reliable in the current five-fold comparison (\texttt{t\ =\ 1.5678}, \texttt{p\ =\ 0.1920}).

This contrast is important because a gain that appeared only on the training set could be dismissed as a consequence of additional flexibility. In the present case, the main advantage appears under cross-validation, where every fold requires generalization to unseen syllogisms. Moreover, the bounded dual-path model is much smaller in size than the direct baseline: the full Experiment 2 model has \texttt{635} trainable parameters, whereas the direct MLP has \texttt{2,505}. This does not prove that the deliberation pathway reasons in the human sense, but it does indicate that the added structure contributes more than memorization of the training split. More broadly, it suggests that a lightweight architectural decomposition can improve performance without requiring a large increase in parameter count, which may make this kind of design attractive for more complex reasoning problems.

The response-type breakdown in Table 4 clarifies where the gain occurs. Relative to intuition, deliberation improves most on \texttt{NVC} (\texttt{+0.2657}), \texttt{Eca} (\texttt{+0.2587}), and \texttt{Oca} (\texttt{+0.1630}), with additional gains on \texttt{Iac} and \texttt{Ica}. These are theoretically important because they concentrate on rejection responses and \texttt{c-a} conclusion directions, which are consistent with classic figural-direction phenomena in the syllogistic literature (Khemlani \& Johnson-Laird, 2012). The result suggests that deliberation is not improving uniformly; it is helping most where premise order, conclusion direction, or the need to reject an initially plausible conclusion matter most. By contrast, \texttt{Aca} remains poorly captured by all three models. In the direct MLP, held-out \texttt{Aca} correlation is slightly negative (\texttt{-0.0170}), whereas the intuition and deliberation pathways are only weakly positive. This likely reflects the rarity and instability of that response category in the human data as well as the general difficulty of \texttt{c-a} universal conclusions.

The item-level pattern points in the same direction. The largest reductions in syllogism-level RMSE occur for \texttt{IA1}, \texttt{EA2}, \texttt{AE2}, \texttt{II1}, \texttt{II4}, \texttt{AI2}, and \texttt{IE1}. Several of these are items with multiple valid conclusions or substantial human mass on \texttt{c-a} responses, and several of the largest invalid-item gains occur on \texttt{NVC}-heavy syllogisms such as \texttt{II1}, \texttt{II4}, and \texttt{IO4}. A fuller one-model versus multiple-model classification is still needed, but the present item-level pattern is already more consistent with structured deliberative sensitivity than with a uniform gain spread across the benchmark.

At the same time, the word \emph{deliberation} should be understood cautiously. The present deliberation pathway is not process-isomorphic to \texttt{mReasoner}: it does not explicitly construct alternative mental models, search sequentially for counterexamples, or revise a symbolic conclusion. The label is intended instead as an outcome-level analogy to a second, more structured stage of computation that improves on a bounded first pass.

The canonical interpretability run uses a fixed \texttt{80:20} train/test split and \texttt{4000} training epochs. On that split, intuition reaches test \texttt{r\ =\ 0.7900} and deliberation reaches test \texttt{r\ =\ 0.8475}, while deliberation also improves RMSE from \texttt{0.1041} to \texttt{0.0946}. We treat this result as illustrative rather than decisive, because the paper's main claims rest on cross-validation. Still, it provides concrete held-out examples for the figures and interpretability analyses that follow. The split contains \texttt{51} training syllogisms and \texttt{13} held-out test syllogisms, so the heatmaps in Figures 4 and 5 should be read as detailed case illustrations rather than as full-benchmark summaries. Figure 3 gives a compact intuition-versus-deliberation performance summary for that canonical split, while Figures 4 and 5 show held-out response matrices for the intuition and deliberation pathways respectively.

The five-seed stability sweep serves a different purpose from the 5-fold cross-validation analysis. Cross-validation provides the main evidence about held-out predictive performance, whereas the repeated \texttt{80:20} runs are used only to assess whether the interpretability patterns recur across different initializations and split memberships. In the current setup, each seed changes both the model's random initialization and the membership of the \texttt{80:20} train/test split. Across the five repeated runs, deliberation test correlation ranges from \texttt{0.8061} to \texttt{0.8739}, and the intuition-to-deliberation gain ranges from \texttt{+0.0111} to \texttt{+0.1688}.

\begin{figure}
\centering
\pandocbounded{\includegraphics[keepaspectratio]{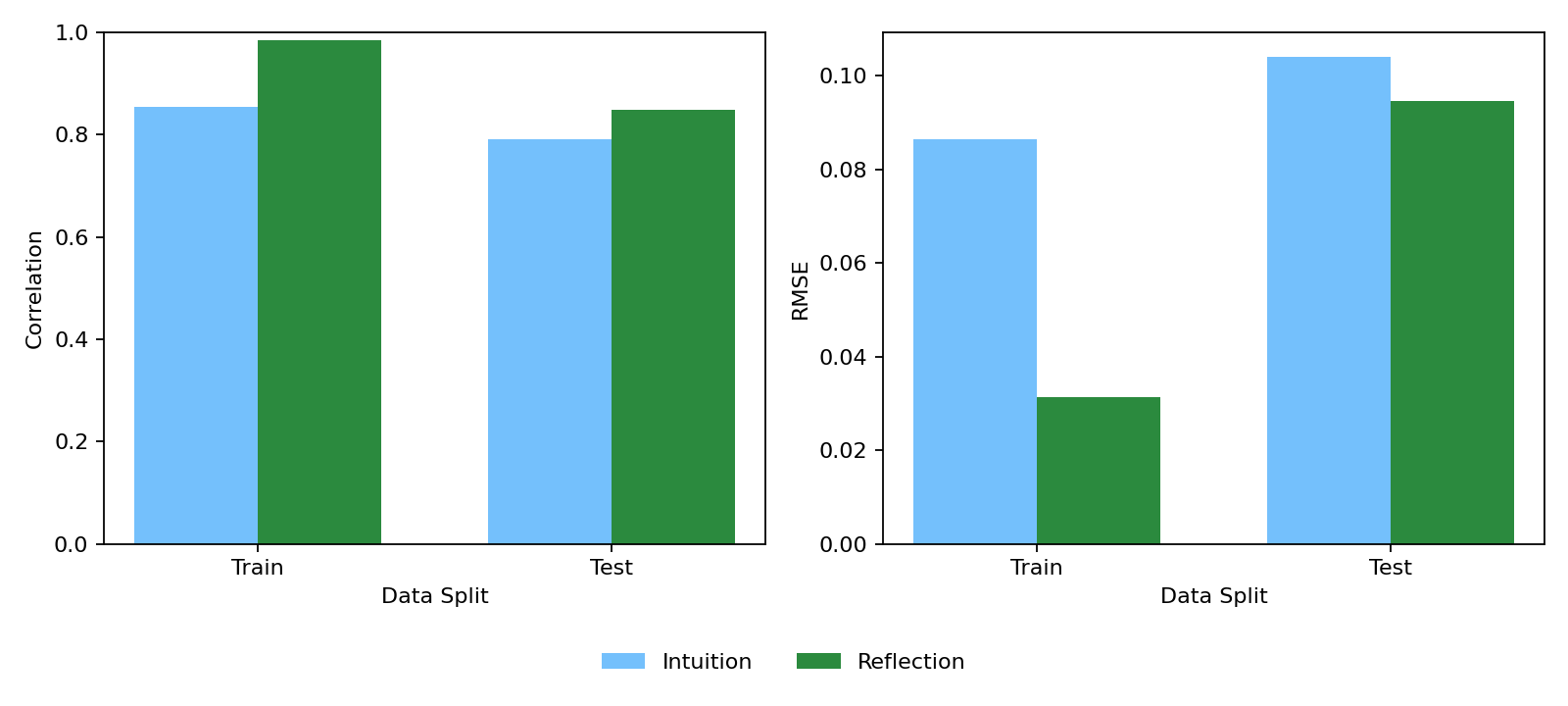}}
\caption{Intuition-versus-deliberation performance on the canonical single split used for interpretability. Deliberation improves both correlation and error relative to intuition.}
\end{figure}

\begin{figure}
\centering
\pandocbounded{\includegraphics[keepaspectratio]{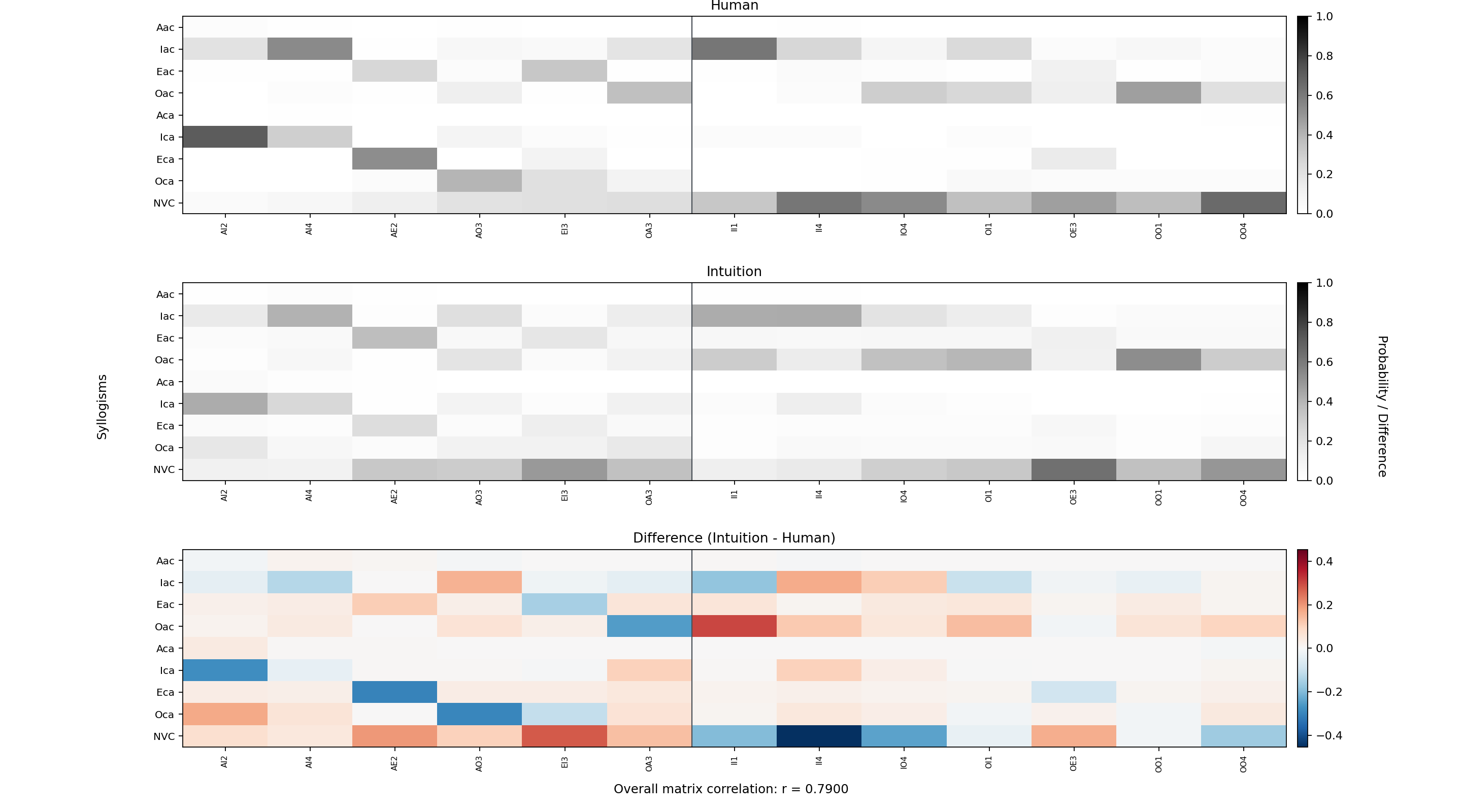}}
\caption{Held-out response-distribution predictions from the bounded intuition pathway in Experiment 2.}
\end{figure}

\begin{figure}
\centering
\pandocbounded{\includegraphics[keepaspectratio]{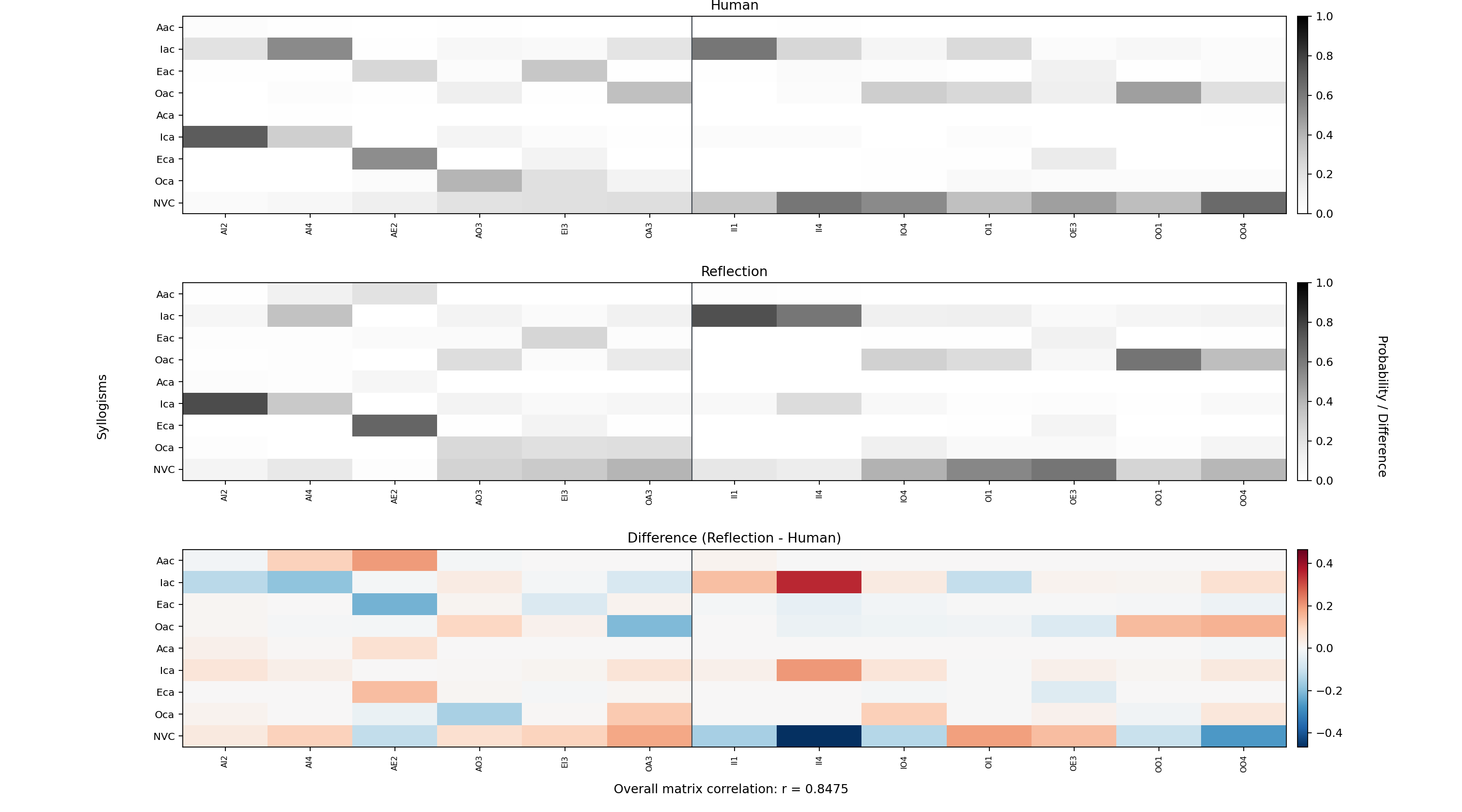}}
\caption{Held-out response-distribution predictions from the deliberation pathway in Experiment 2. Relative to intuition, deliberation more closely tracks the human response matrix on the same held-out items.}
\end{figure}

Taken together, Experiments 1 and 2 support a limited but important conclusion: a bounded deliberation pathway improves over bounded intuition on this syllogistic benchmark, and the gain survives held-out evaluation. That is a stronger result than a simple in-sample fit, and it is the main reason Experiment 2 is more interesting than a generic larger network.

\section{Internal Differentiation and Ablation}\label{internal-differentiation-and-ablation}

The next question is whether the deliberation pathway behaves as a single opaque block or whether it develops differentiated internal structure. To address this question, we first analyze the canonical \texttt{80:20} single-split run of Experiment 2 using activation heatmaps, gate summaries, and inference-time ablation. We then compare those patterns against a five-seed stability sweep over repeated \texttt{80:20} runs. The heatmaps and detailed tables below come from the canonical interpretability split, while the stability sweep is used to determine which aspects of that picture recur across different initializations and different train/test memberships.

\subsection{Specialization patterns}\label{specialization-patterns}

The five deliberative states are not used equally. Table 7 summarizes the five-seed stability sweep, and Table 8 reports the gate-winner counts on the training and test partitions of the canonical split. Here, a gate winner is the state with the highest gate weight for a given syllogism.

\begin{table}[H]
\centering
\small
\caption{Summary of the five-seed stability sweep over repeated \texttt{80:20} runs.}
\begin{tabular}{@{}
  >{\raggedright\arraybackslash}p{(\linewidth - 14\tabcolsep) * \real{0.1111}}
  >{\raggedleft\arraybackslash}p{(\linewidth - 14\tabcolsep) * \real{0.1481}}
  >{\raggedleft\arraybackslash}p{(\linewidth - 14\tabcolsep) * \real{0.1481}}
  >{\raggedleft\arraybackslash}p{(\linewidth - 14\tabcolsep) * \real{0.1481}}
  >{\raggedright\arraybackslash}p{(\linewidth - 14\tabcolsep) * \real{0.1111}}
  >{\raggedright\arraybackslash}p{(\linewidth - 14\tabcolsep) * \real{0.1111}}
  >{\raggedright\arraybackslash}p{(\linewidth - 14\tabcolsep) * \real{0.1111}}
  >{\raggedright\arraybackslash}p{(\linewidth - 14\tabcolsep) * \real{0.1111}}@{}}
\toprule\noalign{}
\begin{minipage}[b]{\linewidth}\raggedright
Seed
\end{minipage} & \begin{minipage}[b]{\linewidth}\raggedleft
Test intuition r
\end{minipage} & \begin{minipage}[b]{\linewidth}\raggedleft
Test deliberation r
\end{minipage} & \begin{minipage}[b]{\linewidth}\raggedleft
Deliberation gain
\end{minipage} & \begin{minipage}[b]{\linewidth}\raggedright
Dominant test state
\end{minipage} & \begin{minipage}[b]{\linewidth}\raggedright
Dead states
\end{minipage} & \begin{minipage}[b]{\linewidth}\raggedright
Oac-leaning state
\end{minipage} & \begin{minipage}[b]{\linewidth}\raggedright
Strongest ablation state
\end{minipage} \\
\midrule\noalign{}
1 & 0.7051 & 0.8739 & +0.1688 & state 2 & state 3, state 5 & state 3 & state 2 \\
2 & 0.7291 & 0.8072 & +0.0781 & state 3 & state 5 & state 3 & state 1 \\
3 & 0.7606 & 0.8141 & +0.0535 & state 1 & state 2, state 4, state 5 & state 5 & state 5 \\
4 & 0.6983 & 0.8061 & +0.1077 & state 1 & state 2, state 3 & state 1 & state 1 \\
5 & 0.8261 & 0.8372 & +0.0111 & state 4 & state 2, state 3, state 5 & state 4 & state 4 \\
\bottomrule
\end{tabular}
\end{table}

Across these five runs, the number of active test-time states ranges from \texttt{2} to \texttt{4}. Here, ``dead states'' refers only to states with zero gate-winner counts on both train and test, not to states with zero total contribution in the soft gate mixture. Every run develops a dominant test-time state and at least one comparatively \texttt{Oac}-leaning state, but the literal state indices vary. In Seed \texttt{1}, for example, \texttt{state\ 3} is identified as the \texttt{Oac}-leaning state and is also listed as dead; this means that it never wins the gate, even though its output distribution, when examined in isolation, is dominated by \texttt{Oac}. Seed \texttt{5} shows the smallest deliberation gain, yet it still exhibits the same sparse role structure, with one dominant test-time state, one \texttt{Oac}-leaning state, and three states that never win the gate.

\begin{table}[H]
\centering
\small
\caption{Gate-winner counts for the five deliberative states in the canonical \texttt{80:20} interpretability split.}
\begin{tabular}{@{}lrr@{}}
\toprule\noalign{}
State & Train gate winners & Test gate winners \\
\midrule\noalign{}
State 1 & 13 & 1 \\
State 2 & 6 & 1 \\
State 3 & 0 & 0 \\
State 4 & 22 & 11 \\
State 5 & 10 & 0 \\
\bottomrule
\end{tabular}
\end{table}

Several states show distinctive behavior. In the canonical split, \texttt{State\ 1} is the clearest specialist. Its average state distribution on the training partition is dominated by \texttt{Oac} at approximately \texttt{0.9918}, and its test distribution is even more concentrated around \texttt{Oac}. \texttt{State\ 3}, by contrast, is effectively unused: it never wins on either train or test. \texttt{State\ 4} acts as the dominant general-purpose deliberative state, especially on held-out items. \texttt{State\ 5} appears important as well, although its profile is broader and less neatly specialized than \texttt{state\ 1}.

If all five states were used equally and interchangeably, it would be natural to describe the deliberative mechanism as a diffuse hidden layer. Instead, the network appears to differentiate among the available deliberative states, assigning some narrow or dominant roles while leaving some states weakly used or effectively idle. The five-seed stability sweep makes this point visible rather than merely asserted: each run develops one dominant test-time state, one especially \texttt{Oac}-leaning state, and several weakly used or unused states, even though the exact state indices change. Figures 6 and 7 visualize the canonical split's internal organization through the deliberation-state activations and gate weights on held-out items. Table 9 summarizes the main specialization claims for the five deliberative states in that canonical case.

\begin{table}[t]
\centering
\footnotesize
\caption{Summary of state specialization in the canonical interpretability split.}
\begin{tabular}{@{}
  >{\raggedright\arraybackslash}p{(\linewidth - 4\tabcolsep) * \real{0.3333}}
  >{\raggedright\arraybackslash}p{(\linewidth - 4\tabcolsep) * \real{0.3333}}
  >{\raggedright\arraybackslash}p{(\linewidth - 4\tabcolsep) * \real{0.3333}}@{}}
\toprule\noalign{}
\begin{minipage}[b]{\linewidth}\raggedright
State
\end{minipage} & \begin{minipage}[b]{\linewidth}\raggedright
Observed role
\end{minipage} & \begin{minipage}[b]{\linewidth}\raggedright
Main evidence
\end{minipage} \\
\midrule\noalign{}
State 1 & \texttt{Oac}-leaning specialist for \texttt{O}/\texttt{E}-minor items & Average state distribution dominated by \texttt{Oac}; removal harms \texttt{Oac} behavior \\
State 2 & Universal-conclusion contributor for \texttt{AA} and \texttt{AE} valid syllogisms & Wins on \texttt{AA1}, \texttt{AA4}, \texttt{AE1}, \texttt{AE3}, and \texttt{AE4}; removal causes moderate held-out loss \\
State 3 & Effectively unused & Never wins on train or test; ablation has almost no effect \\
State 4 & Dominant general-purpose state for many \texttt{I}/\texttt{O}-premise and \texttt{NVC}-heavy items & Most frequent gate winner on train and test; strongest association with \texttt{Iac}, \texttt{Ica}, \texttt{Oca}, and \texttt{NVC}; largest ablation effect \\
State 5 & \texttt{E}-premise / \texttt{EE}-\texttt{EA} contributor & Wins on \texttt{EA*}, \texttt{EI*}, and all four \texttt{EE} items in training; ablation still causes large performance loss \\
\bottomrule
\end{tabular}
\end{table}

\begin{samepage}
\noindent\begin{minipage}[t]{0.48\linewidth}
\centering
\includegraphics[width=\linewidth,keepaspectratio]{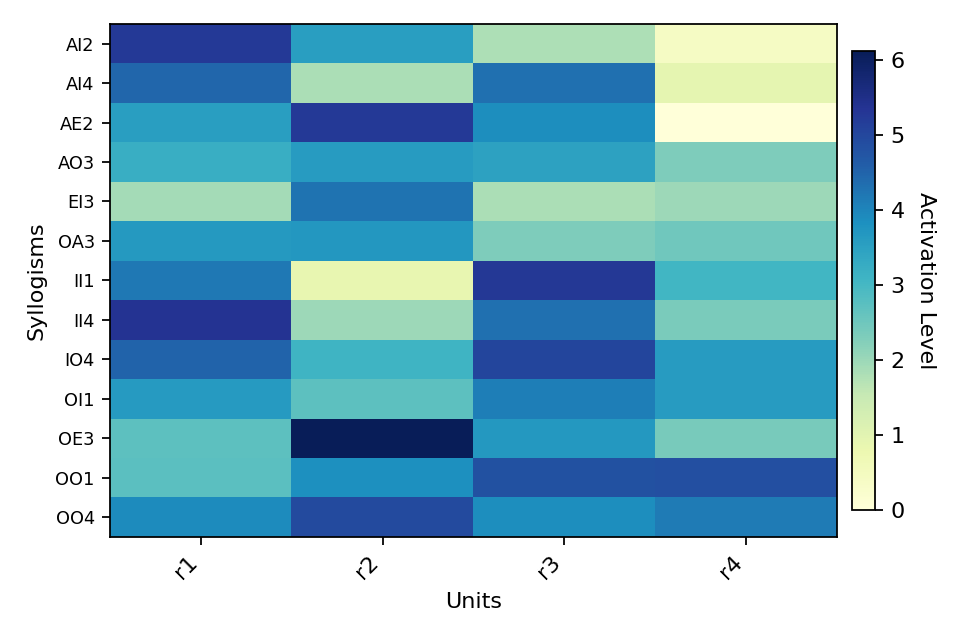}
\captionof{figure}{Held-out deliberation-state activations for the canonical split. Different syllogisms recruit different deliberative patterns rather than a uniform hidden code.}
\end{minipage}\hfill
\begin{minipage}[t]{0.48\linewidth}
\centering
\includegraphics[width=\linewidth,keepaspectratio]{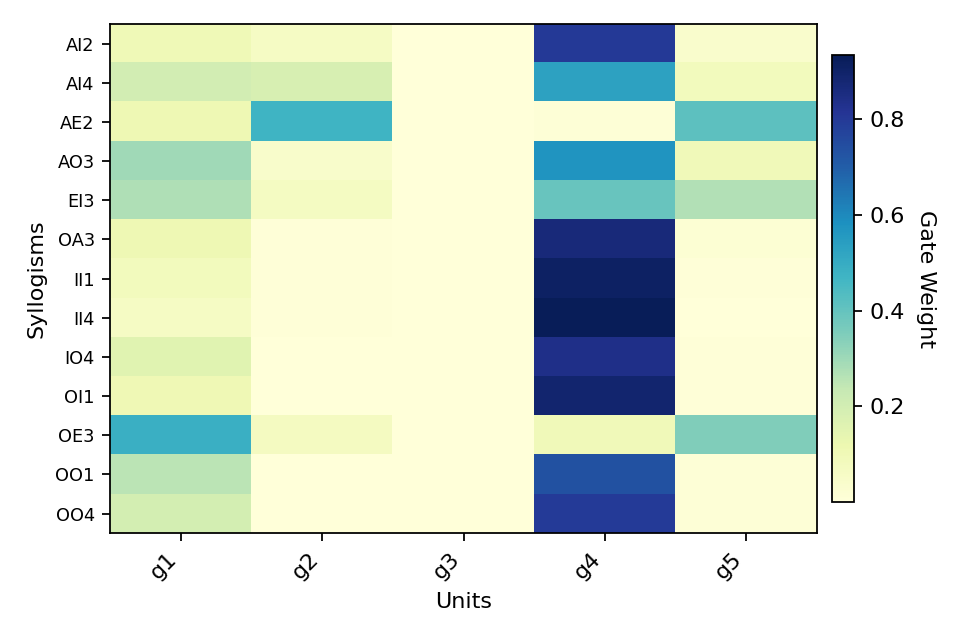}
\captionof{figure}{Held-out gate weights for the five deliberative states. The deliberative mechanism does not use all five states equally, and the broader stability sweep shows that this sparse allocation pattern recurs even when the exact state index assignments change.}
\end{minipage}
\end{samepage}

\subsection{Inference-time ablation}\label{inference-time-ablation}

Specialization patterns are more persuasive when they survive ablation. We therefore remove one deliberative state at a time \emph{after training}, renormalize the remaining gate weights, and recompute the deliberation output at inference time. No retraining occurs during this procedure. Table 10 shows the ablation results for inference on the held-out test partition of the canonical single-split run, not a new optimization run. On the test split, baseline deliberation performance is \texttt{r\ =\ 0.8475} with RMSE \texttt{0.0946}.

\begin{table}[H]
\centering
\small
\caption{Inference-time ablation results on the held-out test partition of the canonical interpretability split.}
\begin{tabular}{@{}
  >{\raggedright\arraybackslash}p{(\linewidth - 8\tabcolsep) * \real{0.1579}}
  >{\raggedleft\arraybackslash}p{(\linewidth - 8\tabcolsep) * \real{0.2105}}
  >{\raggedleft\arraybackslash}p{(\linewidth - 8\tabcolsep) * \real{0.2105}}
  >{\raggedleft\arraybackslash}p{(\linewidth - 8\tabcolsep) * \real{0.2105}}
  >{\raggedleft\arraybackslash}p{(\linewidth - 8\tabcolsep) * \real{0.2105}}@{}}
\toprule\noalign{}
\begin{minipage}[b]{\linewidth}\raggedright
Ablation
\end{minipage} & \begin{minipage}[b]{\linewidth}\raggedleft
Test correlation
\end{minipage} & \begin{minipage}[b]{\linewidth}\raggedleft
Correlation drop
\end{minipage} & \begin{minipage}[b]{\linewidth}\raggedleft
Test RMSE
\end{minipage} & \begin{minipage}[b]{\linewidth}\raggedleft
RMSE increase
\end{minipage} \\
\midrule\noalign{}
none & 0.8475 & 0.0000 & 0.0946 & 0.0000 \\
remove state 1 & 0.7331 & 0.1143 & 0.1310 & 0.0364 \\
remove state 2 & 0.7384 & 0.1091 & 0.1292 & 0.0345 \\
remove state 3 & 0.8474 & 0.0000 & 0.0946 & 0.0000 \\
remove state 4 & 0.2642 & 0.5833 & 0.2700 & 0.1754 \\
remove state 5 & 0.5519 & 0.2956 & 0.1796 & 0.0850 \\
\bottomrule
\end{tabular}
\end{table}

These ablations support three points.

First, \texttt{state\ 3} is genuinely dispensable. Its removal changes essentially nothing. This reinforces the interpretation that it is not just weakly used, but functionally redundant in the trained model.

Second, \texttt{state\ 1} is not merely superficially associated with \texttt{Oac}. Removing it drops overall test correlation to \texttt{0.7331} and increases \texttt{Oac} RMSE from approximately \texttt{0.0903} to \texttt{0.1754}. This is evidence of functional specialization rather than cosmetic clustering.

Third, \texttt{state\ 4} is the dominant workhorse. Removing it causes the largest collapse in held-out performance, dropping test correlation to \texttt{0.2642}. \texttt{State\ 5} also matters substantially, though less dramatically, and \texttt{state\ 2} contributes measurably as well.

\section{Discussion}\label{discussion}

Taken together, the results of this study indicate that a bounded neural architecture can support a meaningful distinction between intuition and deliberation on the syllogistic benchmark used here. Under matched cross-validation, the deliberation pathway improves on bounded intuition, especially on rejection responses and \texttt{c-a} conclusions, and the interpretability analyses suggest that this advantage is not simply the result of a uniform increase in hidden capacity. Instead, the deliberation pathway repeatedly settles into a sparse allocation in which only a subset of the five states carry most of the computation.

One plausible interpretation is that the syllogistic task space itself contains broad families that reward different computational emphases. Some items invite rapid first-pass conclusions, whereas others place more weight on rejecting a tempting conclusion, reversing conclusion direction, or preserving multiple possibilities. In that setting, a learned gate can benefit from allocating different parts of the deliberation mechanism to different regions of the task space. The recurring \texttt{Oac}-leaning state and the recurring dominant workhorse state are consistent with that possibility, even though the exact state indices vary across runs.

The response-type profile is also theoretically suggestive. Gains concentrate on rejection responses and \texttt{c-a} conclusions, which are exactly the parts of the syllogistic response space where a model may need to preserve multiple possibilities rather than commit too quickly to a first-pass conclusion. In that sense, the response-type pattern is consistent with a second-stage computation that is more sensitive to alternative possibilities and directional asymmetries than the bounded intuition pathway. However, the present paper does not yet provide a direct breakdown by syllogistic figure, so any figural interpretation should be understood as suggestive rather than conclusive.

The aggregate bootstrap intervals are also consistent with this descriptive pattern: the deliberation pathway's \texttt{95\%} interval \texttt{{[}0.7608,\ 0.8634{]}} lies above the direct MLP's point estimate of \texttt{0.7105}, although the paper continues to treat the paired fold-wise test as the main inferential result.

The interpretability results strengthen that story again. A skeptical reader could object that the system is only learning distributed statistical associations. However, the present results make that objection less complete than it would be for a standard black-box baseline. The deliberation pathway does not simply use all its components uniformly. Instead, it differentiates among them, allocates one state to \texttt{Oac}-like behavior, shows broader sensitivity to families such as \texttt{AA}/\texttt{AE}, \texttt{EE}/\texttt{EA}, and many \texttt{I}/\texttt{O}-premise items in the canonical split, and repeatedly collapses onto a sparse allocation in which only a subset of the five states carry most of the computation. This pattern hints that the gated architecture is discovering a task decomposition that is sensitive to broad regularities in the syllogistic space rather than only to isolated item-level associations.

These observations do not justify the claim that the deliberative states are explicit symbolic rules in the sense of a hand-written reasoning program. They also do not show that the model deliberates in the process-level sense implemented in \texttt{mReasoner}, where a reasoner can construct a model, search for alternatives, and revise an initial answer. These observations support a narrower but still meaningful claim: the deliberation pathway contains functionally differentiated subcomponents, some of which exhibit distinctive response-family behavior and some of which are causally important at inference time. On that basis, the architecture is best described as showing \emph{reasoning-like internal structure} rather than as having fully recovered human reasoning rules.

It is important to note that the direct MLP baseline remains competitive, and the deliberation pathway's advantage over that baseline is not statistically secure under the current five-fold test. The strongest result is deliberation over intuition, not deliberation over all other models. The direct MLP therefore remains an important comparison point: it shows what a straightforward distribution learner can achieve on the same benchmark without the added architectural structure. Likewise, the internal blocks are functionally differentiated, but they are not yet transparent enough to be identified as explicit mental-model operations in the way \texttt{mReasoner}'s rules and search procedures are. Across the five-seed stability sweep, the same broad functional roles recur even though they are not consistently assigned to the same state indices. The most defensible interpretation is therefore that the model develops stable role structure under a sparse gating regime rather than immutable identities for \texttt{state\ 1}, \texttt{state\ 2}, and so on.

In that sense, the present paper is best understood as a bridge rather than a replacement for computational mental-model theory. It shows that under bounded conditions, a neural architecture can develop a robust deliberation advantage and differentiated deliberative subcomponents on a classic reasoning benchmark. It does not yet show that those subcomponents align neatly with \texttt{mReasoner}'s one-model versus multiple-model distinction, nor does it yet compare its predictions item by item against \texttt{mReasoner}. However, it does identify exactly where the next theoretical bridge should be built: on the response families and item classes for which deliberation helps most. The two most important next analyses are therefore a direct item-by-item comparison with \texttt{mReasoner} and an explicit one-model versus multiple-model breakdown of the 64 syllogisms.

For computer science readers, the main takeaway is that a deliberately bounded learned architecture can outperform a single-pass baseline while developing sparse, functionally differentiated internal subcomponents, and it does so in a regime where the entire internal structure remains legible. The present results establish, in a fully controlled setting, the architectural principle that learned multi-stage computation with sparse gating improves held-out generalization over single-pass baselines. The natural next question is whether that principle, and the interpretability that accompanies it, survive as the input representation becomes less hand-crafted, the task space grows beyond exhaustive enumeration, and the system must learn to extract its own premise structure from richer input. In that sense, the present benchmark serves not only as a micro-testbed for cognitive theory but also as an interpretability anchor point for a research program that may eventually address broader questions about learned reasoning and world models.

\section{Limitations}\label{limitations}

Several limitations are important.

First, the benchmark uses aggregate human response distributions rather than participant-level process data. This means that the models are trained to match population-level outputs, not the sequence, timing, confidence, or revision behavior of individual reasoners. That matters because prior work suggests that syllogistic reasoners differ substantially in their strategies and levels of deliberation (Bacon et al., 2003; Bucciarelli \& Johnson-Laird, 1999).

Second, the scope is restricted to classic syllogisms. Although syllogisms are theoretically important, they are still a narrow domain. Broader claims about reasoning would require transfer to additional tasks and, ideally, tasks for which deliberation is behaviorally necessary rather than simply architecturally available.

Third, the present interpretability results are still limited. Activation heatmaps, gate summaries, ablations, and the five-seed stability sweep provide evidence of internal differentiation. However, these analyses still do not yield a clean symbolic interpretation of each deliberative state. The paper also does not include a sensitivity analysis of the equal \texttt{0.5/0.5} weighting used to combine the intuition and deliberation losses in Experiment 2, so the robustness of the present results to alternative loss-weight choices remains to be tested.

Fourth, the statistical comparison between the deliberation pathway and the direct MLP baseline is not significant under the current five-fold paired test. The strongest robust claim is therefore deliberation over bounded intuition, not deliberation over every alternative.

Finally, none of the present results proves that the model reasons in the full human sense. The most we claim is that the model exhibits bounded deliberation, held-out gains, and internal differentiation consistent with reasoning-like structure. A direct comparison with \texttt{mReasoner} and a one-model versus multiple-model analysis would strengthen the theoretical bridge further.

\section{Conclusion}\label{conclusion}

This paper presented two experiments on neural modeling of syllogistic reasoning. Experiment 1 established a direct baseline for predicting full human response distributions under held-out evaluation. Experiment 2 introduced a bounded dual-path model with separate intuition and deliberation pathways. Under 5-fold cross-validation, deliberation significantly outperformed bounded intuition. Interpretability analyses further showed that the deliberation pathway develops differentiated internal structure, including an \texttt{Oac}-leaning state, a dominant workhorse state, and a sparse allocation in which several states are weakly used or unused, even though the exact state indices vary across runs.

These results do not establish that the model reasons in a fully human-like way. However, they do provide evidence that a bounded neural architecture can move beyond simple in-sample mapping and toward a more structured division of labor between intuition and deliberation. For cognitive science, the results provide a basis for further work on neural models that engage more directly with theories of reasoning, beginning with a direct item-by-item comparison with \texttt{mReasoner} and an explicit one-model versus multiple-model analysis. For AI more broadly, the results suggest a research trajectory in which the same architectural principles are tested under progressively relaxed constraints: less pre-structured input representations, larger and open-ended task spaces, and eventually perceptual grounding. The present paper is the first step in that trajectory, chosen because it is small enough to be fully understood.

\section{Data availability}\label{data-availability}

Raw data are available at OSF: https://osf.io/8zm5u

\section{References}\label{references}

Bacon, A. M., Handley, S. J., \& Newstead, S. E. (2003). Individual differences in strategies for syllogistic reasoning. \emph{Thinking \& Reasoning, 9}(2), 133-168. \url{https://doi.org/10.1080/13546780343000196}

Bucciarelli, M., \& Johnson-Laird, P. N. (1999). Strategies in syllogistic reasoning. \emph{Cognitive Science, 23}(3), 247-303. \url{https://doi.org/10.1016/S0364-0213(99)00008-7}

Cassimatis, N. L., Bello, P., \& Langley, P. (2008). Ability, breadth, and parsimony in computational models of higher-order cognition. \emph{Cognitive Science, 32}(8), 1304-1322. \url{https://doi.org/10.1080/03640210802455175}

Chater, N., \& Oaksford, M. (1999). The probability heuristics model of syllogistic reasoning. \emph{Cognitive Psychology, 38}(2), 191-258. \url{https://doi.org/10.1006/cogp.1998.0696}

De Neys, W. (2014). Conflict detection, dual processes, and logical intuitions: Some clarifications. \emph{Thinking \& Reasoning, 20}(2), 169-187. \url{https://doi.org/10.1080/13546783.2013.854725}

De Neys, W. (2023). Advancing theorizing about fast-and-slow thinking. \emph{Behavioral and Brain Sciences, 46}, Article e111. \url{https://doi.org/10.1017/S0140525X2200142X}

Evans, J. St.~B. T. (2008). Dual-processing accounts of reasoning, judgment, and social cognition. \emph{Annual Review of Psychology, 59}, 255-278. \url{https://doi.org/10.1146/annurev.psych.59.103006.093629}

Hattori, M. (2016). Probabilistic representation in syllogistic reasoning: A theory to integrate mental models and heuristics. \emph{Cognition, 157}, 296-320. \url{https://doi.org/10.1016/j.cognition.2016.09.009}

Johnson-Laird, P. N., \& Ragni, M. (2024). Reasoning about possibilities: Modal logics, possible worlds, and mental models. \emph{Psychonomic Bulletin \& Review, 31}, 2143-2178. \url{https://doi.org/10.3758/s13423-024-02518-z}

Johnson-Laird, P. N., \& Steedman, M. (1978). The psychology of syllogisms. \emph{Cognitive Psychology, 10}(1), 64-99. \url{https://doi.org/10.1016/0010-0285(78)90019-1}

Khemlani, S., \& Johnson-Laird, P. N. (2012). Theories of the syllogism: A meta-analysis. \emph{Psychological Bulletin, 138}(3), 427-457. \url{https://doi.org/10.1037/a0026841}

Khemlani, S., \& Johnson-Laird, P. N. (2022). Reasoning about properties: A computational theory. \emph{Psychological Review, 129}(2), 289-312. \url{https://doi.org/10.1037/rev0000240}

Tessler, M. H., Tenenbaum, J. B., \& Goodman, N. D. (2022). Logic, probability, and pragmatics in syllogistic reasoning. \emph{Topics in Cognitive Science, 14}(3), 574-601. \url{https://doi.org/10.1111/tops.12593}